\documentclass[twoside]{article}

\usepackage{appendix}

\usepackage{amsmath,amsfonts,bm}

\def\1{\bm{1}}

\def\vh{{\bm{h}}}

\def\vx{{\bm{x}}}

\def\mB{{\bm{B}}}

\def\mW{{\bm{W}}}
\def\mX{{\bm{X}}}

\DeclareMathAlphabet{\mathsfit}{\encodingdefault}{\sfdefault}{m}{sl}
\SetMathAlphabet{\mathsfit}{bold}{\encodingdefault}{\sfdefault}{bx}{n}

\def\gG{{\mathcal{G}}}

\def\gV{{\mathcal{V}}}

\usepackage{natbib}
\setcitestyle{authoryear,open={(},close={)}} 

\usepackage{url, bm}

\usepackage{multirow}

\usepackage{mathtools}

\usepackage{floatrow}
\newfloatcommand{capbtabbox}{table}[][\FBwidth]

\usepackage{amsmath,amsfonts,amssymb}

\usepackage{microtype}

\usepackage{algorithm}
\usepackage{algorithmic}

\usepackage{subfigure}
\usepackage{diagbox}
\usepackage{adjustbox}
\usepackage{booktabs}
\usepackage{makecell}
\usepackage{tabularx}
\usepackage{multirow}
\usepackage{tabu}
\usepackage{footnote}
\usepackage{subfigure}
\usepackage{caption}
\usepackage{xcolor}

\newcommand{\MI}{\mathcal{I}}
\newcommand{\SP}[1]{J({#1})}
\newcommand{\SPs}{J}
\newcommand{\genSent}{S}
\newcommand{\genSentP}{\hat S}
\newcommand{\natSent}{A}
\newcommand{\natSentP}{\hat{A}}
\newcommand{\Hid}[1]{H({#1})}
\newcommand{\hidt}[1]{h_t({#1})}
\newcommand{\Hids}{H}
\newcommand{\hidts}{h_t}

\usepackage[accepted]{aistats2023}
\begin{document}

\runningtitle{Toward Fairness in Text Generation via Mutual Information Minimization based on Importance Sampling}

%

%

\twocolumn[

\aistatstitle{Toward Fairness in Text Generation via Mutual Information Minimization \\ based on Importance Sampling}

\aistatsauthor{ Rui Wang$^*$ \And Pengyu Cheng$^*$ \And  Ricardo Henao$^\dagger$ }


\aistatsaddress{ Duke University \And  Tencent AI Lab \vspace{3mm}\\ \texttt{rui.wang16@duke.edu}  \And Duke University $\&$  KAUST } ]

 

\begin{abstract}
Pretrained language models (PLMs), such as GPT-2, have achieved remarkable empirical performance in text generation tasks. 
However, pretrained on large-scale natural language corpora, the generated text  from PLMs may exhibit \textit{social bias} against disadvantaged demographic groups.
%
%
To improve the fairness of PLMs in text generation, we propose to
minimize the mutual information between the semantics in the generated text sentences and their \textit{demographic polarity}, \emph{i.e.}, the demographic group to which the sentence is referring.
In this way, the mentioning of a demographic group (\emph{e.g.}, \textit{male or female}) is encouraged to be independent from how it is described in the generated text, thus effectively alleviating the social bias.
Moreover, we propose to efficiently estimate the upper bound of the above mutual information via importance sampling, leveraging a natural language corpus.
We also propose a distillation mechanism that preserves the language modeling ability of the PLMs after debiasing.
%
%
Empirical results on real-world benchmarks demonstrate that 
the proposed method yields superior performance  in term of both fairness and language modeling ability.
\end{abstract}

\begin{table*}[!t]
\centering
\resizebox{\columnwidth}{!}{%
  \begin{tabular}{c|c} \hline 
  Types & Sentences  \\ \hline
  \multirow{3}{*}{\begin{tabular}{@{}c@{}} Polarity: \textit{Male} \\ (Polarized) \end{tabular}}  & \textbf{He} can make a horse, but \textbf{he} can't make a pony. \textbf{He} will always try to buy things. \\
                                     & \textbf{He} writes at a snail's pace, writes in a high-pitched tone and sometimes turns the page.\\
                                 & a \textbf{man} who is willing to sacrifice \textbf{himself} for \textbf{his} beloved, for \textbf{his} neighbor and for \textbf{his} \textbf{father}\\ \hline
  \multirow{3}{*}{\begin{tabular}{@{}c@{}}Polarity:\textit{Female} \\ (Polarized) \end{tabular}}  & \textbf{she} is ready for \textbf{her} nursing home needs.\\
                                     & The anesthesiologist assistant works under the medical ills of \textbf{her} job. \\
                                 & The first psychiatric nurses faced difficult working ills during \textbf{her} first year at a clinic.\\ \hline
  \multirow{3}{*}{\begin{tabular}{@{}c@{}}\textit{Neutral} \\ (Non-Polarized) \end{tabular}}  & the clinician may use \textbf{his} or \textbf{her} abilities and abilities as an expert in the specific treatment. \\
                                     & Anesthesiologist assistants work in all facets of ills such as headache \\
                                 & Chaplains also build relationships with students  to help them learn who they're talking to.\\  \hline
  \end{tabular}
}
\caption{Examples of sentences generated from the GPT-2 grouped according to whether the sentence is neutral or polarized in terms of gender, \emph{i.e.}, \textit{male} or \textit{female}. The demographic-sensitive words are marked with bold font. The "Polarity" is short for demographic polarity.\vspace{-5mm}} \label{tb: polarity}

\end{table*}

\section{INTRODUCTION} \label{sc: intro}
The recent advent of Pretrained Language Models (PLMs), \emph{e.g.}, GPT-2~\citep{radford2019language}, has tremendously advanced the state-of-the-art for natural language generation tasks~\citep{adiwardana2020towards, yao2019plan, dong2021survey}.
These advances have resulted in human-like cohesive text generation when prompted with a sequence of context words. 
However, existing PLMs are generally pretrained with large-scale natural language corpora crawled from the internet~\citep{schick2021self}, without attentive filtering or scrutiny about the potential undesirable \textit{social bias} exhibited in human language, \emph{i.e.} prejudices or stereotypes  against disadvantaged demographic groups in terms of, \emph{e.g.}, genders or religions. 
Consequently, such a bias may be inherited or worse, exacerbated by the resulting PLMs during pretraining~\citep{barikeri2021redditbias, basta2019evaluating, zhao2019gender}.
For example, recent studies~\citep{dhamala2021bold,lauscher2021sustainable} have shown that the pretrained GPT-2 model can spuriously correlate  a demographic group, \emph{e.g.}, ``\textit{male}'' or ``\textit{female}'', with certain occupations.
For instance, the model can be prone to generate ``\textit{doctor}'' given ``\textit{He works as}'', while tending to generate ``\textit{nurse}'' given ``\textit{She works as}''.
Such stereotypical inclinations in the text generation may cause a negative ethical impact in socio-technical scenarios~\citep{lauscher2021sustainable}, which severely curtails the applicability of PLMs to real-world text generation.

Several methods have been proposed to alleviate social bias for fair text generation. In one direction, these works debias the PLMs from the data perspective. \citet{zhao2018learning} proposed the Conterfactual Data Augmentation (CDA), which creates counterfactual text instances that counter the social bias in the training corpus, \emph{e.g.}, for gender bias, by replacing ``\textit{She}'' with ``\textit{He}'', and {\em vice versa}.
The PLMs are finetuned on both the original corpus and the counterfactual instances.
However, \citet{dinan2019queens} showed that these artificial instances may not be grammatically correct. 
Instead, they manually collect data that is deemed to be free from social bias. 
Unfortunately, such a process is generally laborious and expensive.
From a different perspective, there are also works focusing on the geometry of the pretrained embedding space of PLMs~\citep{bolukbasi2016man, liang2021towards}.
Specifically, they project the (contextualized) token embeddings into the orthogonal space of a demographic biasing subspace, which is linearly spanned by a number of bias directions in the embedding space.
The projection is generally implemented via minimizing the cosine similarity between the embedding vectors and the bias directions~\citep{liang2021towards}.
However, these approaches make a strong assumption about the linearity of the bias in embedding space.
Moreover, there is no guarantee that the cosine similarity can sufficiently capture the bias degree of the learnt embeddings against different demographic groups.
In \citet{kurita2019measuring}, it is shown that cosine similarity based methods do not produce consistent results in measuring bias for token embeddings.
Recently, \citet{henlein2022toothbrushes} showed that the information captured by different handcrafted similarity metrics on pretrained token embeddings, \emph{e.g.}, cosine or Euclidean similarities, widely varies and may not be fully explainable.


Different from previous works, our proposed method does not rely on manual data collection nor handcrafted similarity metrics in the pretrained embedding space.
Alternatively, we propose to debias the PLMs for text generation via minimizing the mutual information between the \textit{demographic polarity} of the generated sentence and its semantics.
The term of \textit{demographic polarity} follows \citet{dhamala2021bold}, denoting which demographic group the sentence is referring to.
For instance, it indicates whether a sentence is referring to a ``\textit{male}'' or a ``\textit{female}'' in a gender debiasing scenario.
By minimizing its mutual information with the sentence semantics, we break the stereotypes regarding demographic groups in text generation such that the mentioning of each demographic group ({\em e.g.}, ``\textit{She}'' or ``\textit{He}'') is decoupled or independent from how it is described in the generated context ({\em e.g.}, ``\textit{Doctor}'' or ``\textit{Nurse}'').
We also propose a distillation mechanism that preserves the language modeling ability of PLMs.
Experiments on real-world benchmarks show that our method outperforms other baselines in terms of both fairness for different demographic groups and language modeling ability.

\section{BACKGROUND}
%
\subsection{Demographic Polarity} \label{sc: sp}
%
Let $\gG = \{G_i\}_{k=1}^K$ be the collection of $K$ demographic groups being considered for fairness in a text generation task. 
As an example, we can have $\gG = \{\textit{male}, \textit{female}\}$ when debiasing for fairness over genders.
Each of such groups, $G_i$, is associated with a set of demographic-sensitive words, $\gV_{G_i}$, indicating how  $G_i$ is manifested in the text, \emph{e.g.}, we can have $\gV_\text{male}=\{\textit{he}, \textit{himself}, \textit{father}, \dots, \emph{etc}.\}$ and $\gV_\text{female}=\{\textit{she}, \textit{herself}, \textit{mother}, \dots, \emph{etc}.\}$.
The \textit{demographic polarity} of a sentence, taking values from $\gG$, is a term that describes which demographic groups the sentence is referring to.
Demographic polarity can be inferred from the mentions of different demographic groups in a sentence. 
Here, we identify the mention of a demographic group $G_i$ as an occurrence of any of its corresponding demographic-sensitive words, $\gV_{G_i}$, in the sentence.
For instance, \citet{dhamala2021bold} quantitatively defines the demographic polarity for gender, {\em i.e.}, gender polarity, as the gender (\textit{male} or \textit{female}) with the highest number of mentions (highest frequency) in a sentence.
Similar to \citet{dhamala2021bold}, we define the demographic polarity of a sentence as the demographic group from $\gG$ with the highest number of mentions. 
Moreover, a sentence is denoted as \textit{polarized} if there exists a  demographic group that is mentioned with higher frequency (higher number of mentions) than the others.
Alternatively, we also denote a sentence as \textit{neutral} or without  demographic polarity, if there is no mentions of any considered demographic group or different  demographic groups have the same number of mentions.
In Table \ref{tb: polarity}, we show examples of sentences generated from GPT-2 that are neutral or polarized in terms of gender, \emph{i.e.}, with $\gG = \{\textit{male}, \textit{female}\}$.

\vspace{-4mm}
\subsection{Language Modeling} \label{sc: model_arc}
\vspace{-1mm}

In this paper, we term a PLM without debiasing for fair text generation, \emph{e.g.}, the pretrained GPT-2, as a \textit{reference} PLM. 
Let $\gV$ be the vocabulary of discrete text tokens.
We define $\mX=(\vx_1, \vx_2, \dots, \vx_T)$ as a sentence of length $T$, where $\vx_t \in \gV$ and $\mX_{<t} = (\vx_1, \vx_2, \dots, \vx_{t-1})$ is the prefix of sentence $\mX$ with length $t-1$, \emph{i.e.}, a partial sentence.
Conditioned on $\mX_{<t}$, the reference model is expected to predict the probability of occurrence for the next token $\vx_t\in \gV$ at position $t$, which is formulated as

\begin{equation} \label{eq:lm-next-token}
    P^\textit{\rm Ref}(\vx_t|\mX_{<t};\gV)= \frac{exp(\,f(\mX_{<t})^\intercal e(\vx_t)\,)}{\sum_{\vx \in \gV}exp(\,f(\mX_{<t})^\intercal e(\vx)\,)} , 
\end{equation}
where $e(\vx)$ is the embedding vector for token $\vx$, and $f(\cdot)$ is the context encoder, both of which have been pretrained on a large-scale text corpus.
For debiasing, we adopt a \textit{post hoc} approach~\citep{cheng2021fairfil, liang2021towards}. Specifically, instead of retraining the whole reference model, we 
stack a trainable debiasing layer $D(\cdot)$ on top of $f(\cdot)$. 
Formally, we write the probability of predicting the next token $\vx_t$ from the debiased PLM as,
\begin{equation} \label{eq:lm-next-token_de}
    \hspace{-2mm}P^{\rm Deb}(\vx_t|\mX_{<t};\gV)\!=\! \frac{exp(\,D\!\circ\! f(\mX_{<t})^\intercal e(\vx_t)\,)}{\sum_{\vx \in \gV}exp(\,D\!\circ\! f(\mX_{<t})^\intercal e(\vx)\,)} , 
\end{equation}
%

where $\circ$ denotes the function composition.
$D$ is implemented as a residual module~\citep{he2016deep}, and is trained for debiasing with a text corpus of natural language.

Let $\genSent$  be the output distribution  over sentences generated from the debiased PLM with $D \circ f(\cdot)$.
The likelihood of sentence $\mX$ with distribution $\genSent$ is
\begin{equation} \label{eq: LM}
    P_\genSent(\mX)= \sigma(\vx_1) \prod_{t=2}^T P^\text{Deb}(\vx_t| \mX_{<t}) ,
\end{equation}
%
%
%
%
where $\sigma(\vx_1)$ is a prior distribution over the start token $\vx_1$ of any sentence $\mX$ in natural language.
Hereafter, we use a subscript to denote the distribution from which $\mX$ is drawn.
Let $\mX$ be a polarized sentence, we define $\genSentP$ as the distribution over polarized sentences  from the debiased reference model, \emph{i.e.}, $P_{\genSentP}(\bm X)= P_{\genSent}(\bm X) /  P_{\genSent}^{\rm polar}$, where $P_{\genSent}^{\rm polar}$ is the probability of generating a polarized sentence from $\genSent$.

\section{DEBIASING PLMs FOR FAIR TEXT GENERATION} \label{sc: debiasing}
As mentioned above, PLMs pretrained with natural language have been shown to manifest social bias in text generation, \emph{i.e.}, by undesirably associating a demographic group with semantics that reflect social stereotypes and prejudices.
Such an association can be understood as a having a skewed likelihood in mentioning different demographic groups (\emph{e.g.}, \textit{male or female}), conditioned on the existence of certain semantics from the context (\emph{e.g.}, \textit{doctor or nurse}), as described in Section \ref{sc: intro}.
In addressing this issue, we propose to minimize mutual information between the demographic polarity of the polarized sentences generated from PLMs and their sentence semantics.
For instance, by minimizing such mutual information for gender debiasing, \emph{i.e.}, $\gG = \{\textit{male}, \textit{female}\}$,
we can encourage that the generation of occupational semantics ({\em e.g.}, ``\textit{doctor}'' or ``\textit{nurse}'') to be independent of whether the sentence is describing a ``\textit{male}'' or ``\textit{female}'', thus effectively alleviating the social bias.

Below, in Section \ref{sc: mi} we explain the proposed debiasing method for fair text generation, and in Section \ref{sc: distill} we introduce a distillation mechanism that preserves the language modeling performance of the debiased PLMs.

\subsection{The Mutual Information Between Demographic Polarity and Sentence Semantics} \label{sc: mi}
%
%
%

For $\mX\sim \genSentP$, the demographic polarity of sentence $\mX$ is denoted  as $\SP{\mX}$.
We represent the sentence semantics of $\bm X$ as the hidden states from context encoder $D \circ f$, which we denote as $\Hid{\mX}$.
Specifically, we have $ H(\mX)=\{\hidt{\mX}\}_{t=1}^T$, where $ \hidt{\mX}={D}\circ f(\mX_{<t})$.
As above, we want to minimize the mutual information between the demographic polarity $\SP{\mX}$ for a polarized sentence $\mX \sim \genSentP$, and its sentence semantics $\Hid{\mX}$.
Their mutual information is formally defined as
\begin{align} \label{eq: mi}
    &\MI(\SP{\mX}, \Hid{\mX}| \bm X \sim \genSentP) = \\
    &\hspace{32mm} \mathbb E_{\bm X \sim \genSentP}  \log \frac{ P_{\genSentP}(\SP{\mX}, \Hid{\mX})}{P_{\genSentP}(\SP{\mX}) P_{\genSentP}(\Hid{\mX})} , \notag
\end{align}


%
where $P_{\genSentP}(\SP{\mX}, \Hid{\mX})$ is the joint distribution over $\SP{\mX}$ and  $\Hid{\mX}$,
while $P_{\genSentP}(\SP{\mX})$ and $P_{\genSentP}(\Hid{\mX})$ are their marginal distributions.
In the following, we  denote $\MI(\SP{\mX}, \Hid{\mX}| \bm X \sim \genSentP) $ as $\MI(\SPs, \Hids) $ for conciseness.


The difficulty in optimizing with $\MI(\SPs, \Hids) $ lies in how to efficiently estimate such mutual information from a text generator. 
For minimization, we first decompose $\MI(\SPs, \Hids) $  over $\Hids$ for each position $t$ as
%
\begin{equation}
    \frac{1}{T} \MI(\SPs, \Hids)  \leq  \frac{1}{T} \sum_{t=1}^{T} \MI(\SPs, \hidts) \label{eq: decom}
\end{equation}
where $\MI(\SPs, \hidts)$ is the mutual information between $\SP{\mX}$ and the position-wise hidden state $\hidt{\mX}$ at position $t$, \emph{i.e.}, replacing $\Hid{\mX}$ with $\hidt{\mX}$ in \eqref{eq: mi}.
$\MI(\SPs, \hidts)$ can be approximated by an existing upper bound estimator of mutual information~\citep{cheng2020club}, defined as
$\MI(\SPs, \hidts) \leq \MI^u(\SPs, \hidts)$, with
%
\begin{align} 
   \MI^u(\SPs, \hidts) = & \ \mathbb E_{\bm X\sim \genSentP} \big(\,  \log P_{\genSentP}(\SP{\mX}|\hidt{\mX})  \notag \\
    & - \mathbb E_{{\bm X^\prime}\sim \genSentP} \log  P_{\genSentP}(\SP{\mX}|\hidt{{{\bm X^\prime}}}) \,\,\big). \label{eq: club}
\end{align}

$\MI^u(\SPs, \hidts)$ is an upper bound that measures the difference between the  log likelihood, $\log P_{\genSentP}(\cdot|\cdot)$, of the positive pair $\{\SP{\mX}, \hidt{\bm X}\}$ 
and the negative pair $\{\SP{\mX}, \hidt{\bm X^\prime}\}$, with $\mX$ and $\mX^\prime$ being independent.
This can be understood as the expected uncertainty of $\SP{\mX}$/$\hidt{\bm X}$ given $\hidt{\bm X}$/$\SP{\mX}$.
Note that we do not follow the Donsker-Varadhan representation~\citep{belghazi2018mutual} since it induces a low bound of mutual information, thus is inappropriate for minimization problems.
From \cite{cheng2020club}, \eqref{eq: club} can be estimated by sampling $\bm X,{\bm X^\prime} \sim \genSentP$,
%
\begin{align} \label{eq: club_est}
   \MI^u(\SPs,\hidts) \approx & \ \frac{1}{N} \sum_{i=1}^{N}  \big(\, q(\SP{\bm X_i}|\hidts(\bm X_i)) \\
   & - \frac{1}{N} \sum_{j=1}^{N} q(\SP{\bm X_i}|\hidts(\bm X'_j))\,\, \big) , \notag
\end{align}
%
where $q(\cdot|\cdot)$ is an approximation to the (unknown) log likelihood $\log P(\cdot|\cdot)$ in \eqref{eq: club}. $q(\cdot|\cdot)$ is trained using samples from the the positive pairs $\{\SP{\mX_i}, \hidt{\bm X_i}\}_{i=1}^N$ in each iteration, as illustrated in Algorithm \ref{alg: overall}.
The sets of samples for $\mX$ and $\mX^\prime$ are denoted as $\{\mX_i\}_{i=1}^N$ and $\{\mX_j^\prime\}_{j=1}^N$, respectively, which are all sampled from $\genSentP$.
For computational efficiency, we let $\bm X$ and ${{\bm X^\prime}}$ share the same set of samples, \emph{i.e.}, $\{\mX_i\}_{i=1}^N=\{\mX_j^\prime\}_{j=1}^N$.

\subsection{Estimating with Natural Language Sentences}
The estimation of the mutual information with \eqref{eq: club_est} requires sampling polarized sentences from $\genSentP$, a common approach for which is to sample from $\genSent$ with the debiased PLM, then only keep those that are polarized~\citep{ dhamala2021bold, martino2010generalized}. 
However, such a sampling strategy for estimating \eqref{eq: club} can be cumbersome for mainly two reasons.
First, $\genSentP$ is constantly changing with the debiased PLM during training, which implies that the samples in \eqref{eq: club_est} have to be regenerated every several iterations.
This reduces the training efficiency by further considering that the generation of words within a sentence, {\em e.g.}, with GPT-2, needs to be done sequentially, thus cannot be parallelized.
Moreover, polarized sentences may only amount to a small portion of the PLM outputs.
For instance, \cite{dhamala2021bold} estimates that only $6.75\%$ sentences generated from GPT-2 are  polarized in term of gender given occupational prompts.\footnote{Also from \citet{liang2021towards}, across 5 natural language corpus, only 1.6\% sentences are gender related and 0.12\% are religion related. Thus, we should not expect high probability in generating polarized sentences from PLMs pretrained over natural language.}
Thus, directly generating $\mX\sim \genSentP$ from the PLMs for estimating \eqref{eq: club} can be extremely inefficient.

In our approach, we estimate \eqref{eq: club} via importance sampling, leveraging a text corpus of natural language.
We describe the natural language corpus used for experiments in Section \ref{sc: experiment}.
Here, we denote $\natSent$ as the distribution over text sentences from natural language, and $\natSentP$ as the conditional distribution of $\natSent$ over polarized sentences.
Let $\mX$ be a polarized sentence, we have $P_{\natSentP}(\bm X)=P_{\natSent}(\bm X)/ P_{\natSent}^{\rm polar} $, where $P_{\natSent}^{\rm polar}$ is the probability of a natural language sentence being polarized.
To circumvent the inefficiency in  sampling from $\genSentP$ with the PLMs outputs (as required in \eqref{eq: club_est}),
we rewrite the upper bound $\MI^u(\SPs, \hidts)$ in \eqref{eq: club} 
in terms of expectation over $\natSentP$, so that we can estimate \eqref{eq: club} via drawing samples from $\natSentP$ instead of $\genSentP$.
Specifically, $\MI^u(\SPs, \hidts)$  can be rewritten as,
%
\begin{align}
     &\MI^u(\SPs, \hidts) = \mathbb E_{\bm X\sim \natSentP} \hat R(\mX) \big(\,  \log P(\SP{\mX}|\hidt{\bm X}) \label{eq: imp} \\ 
     &\hspace{6mm} -   \mathbb E_{{\bm X^\prime}\sim \natSentP}  \hat R(\mX^\prime) \log  P(\SP{\mX}|\hidt{{\bm X^\prime}})\,\, \big), \notag 
\end{align} 
%
%
where $\hat R(\cdot)={P_{\genSentP}({\cdot)}}/{P_{\natSentP}({\cdot})}$ denotes the likelihood ratio.
Note that \eqref{eq: imp} enable us to estimate $\MI^u(\SPs, \hidts)$ via sampling from $\natSentP$, which facilities the training efficiency from two perspectives: \textit{i)} 
unlike $\genSentP$, the distribution $\natSentP$ can be sampled efficiently by conditioning, {\em i.e.}, via filtering out non-polarized sentences from the natural language corpus; and
\textit{ii)} 
the sampled sentences can be reused for estimating \eqref{eq: club} during different steps of training.
%
%
Consequently, following \eqref{eq: club_est}, we estimate \eqref{eq: imp} with sampled sentences from $\natSentP$ using,
\begin{align}
    \MI^u(\SPs, \hidts) \approx & \ \frac{1}{N} \sum_{i=1}^{N} \hat R(\mX_i) \bigl( \,\, q(\SP{\mX_i}|\hidt{\bm X_i}) \label{eq: imp_est} \\
      & - \frac{1}{N} \sum_{j=1}^{N} \hat R(\mX_j)  q(\SP{\mX_i}|\hidt{{\bm X'_j}})\,\, \bigr). \notag
\end{align}


%
%
Similar to \eqref{eq: club_est}, $\bm X$ and ${\bm{ X^\prime}}$ in \eqref{eq: imp} share the same set of samples, \emph{i.e.}, $\{\mX_i\}_{i=1}^N=\{\mX_j^\prime\}_{j=1}^N$ in \eqref{eq: imp_est}.
Note that $\{\mX_i\}_{i=1}^N$ and $\{\mX_j^\prime\}_{j=1}^N$ in \eqref{eq: imp_est} are sampled from $\natSentP$ instead of $\genSentP$.
%
%

The remaining question is how to efficiently calculate the likelihood ratio terms in \eqref{eq: imp_est}.
We address this by leveraging the pretrained knowledge from PLMs.
Specifically, we propose to quantify the probability distribution of $\natSentP$ over natural language sentences using the reference PLM, since it is pretrained with large-scale natural language corpus.
Recall in \eqref{eq:lm-next-token},  $P^{\rm Ref}(\vx_t|\mX_{<t};\gV)$ is the reference probability of generating the next token $\vx_t$ given its context $\mX_{<t}$.
Following \eqref{eq: LM}, we can compute $P_{\natSent}(\bm X)$ as,
\begin{align} \label{eq: LM_gpt}
    P_{\natSent}(\bm X)=\sigma(\vx_1) \prod_{t=2}^T P^{\rm Ref}(\vx^t|\mX_{<t};\gV) ,
\end{align}
%
%
In the experiments, we use the pretrained GPT-2 model~\citep{radford2019language} as our reference PLM.
With \eqref{eq: LM_gpt}, $\hat R(\mX)$ is calculated as,
\begin{align}
   \hat R(\mX) &= \frac{P_{\genSentP}(\bm X)}{P_{\natSentP}(\bm X)} = \underbrace{ \frac{P_{\genSent}(\bm X)}{P_{\natSent}(\bm X)}}_{l(\mX)}
   \cdot
   \underbrace{\frac{P_{\natSent}^{\rm polar}}{P_{\genSent}^{\rm polar}}}_{R^{\rm polar}} \label{eq: lh_ratio} \\
   & = \frac{\prod_{t>1} P^{\rm Deb}(\vx^t|\mX_{<t};\gV))}{\prod_{t>1} P^{\rm Ref}(\vx^t|\mX_{<t};\gV)) }
   \cdot
   R^{\rm polar}\label{eq: lh_ratio_1}
\end{align}
%
%

%
where $P_{\genSent}(\bm X)$ and $P_{\natSent}(\mX)$ are defined in \eqref{eq: LM} and \eqref{eq: LM_gpt}, respectively.
From \eqref{eq: lh_ratio} to \eqref{eq: lh_ratio_1}, $\sigma(\vx_1)$ in $P_{\genSent}(\bm X)$ and $P_{\natSent}(\mX)$ cancels out in the first term $l(\mX)$.
Thus, $l(\mX)$ can be directly calculated with the reference and debiased PLMs using the first term in \eqref{eq: lh_ratio_1}.
However, we note that the value of $R^{\rm polar}$ in \eqref{eq: lh_ratio_1} is not readily available.
Further, since $P_{\genSent}^{\rm polar}$ changes along with training of the debiased PLM, the value of $R^{\rm polar}$ is not a constant during training, thus cannot be ignored in minimization.
To solve this, we approximate $R^{polar}$ with the sampled sentences from $\natSentP$ by noting that,
\begin{align}
\frac{1}{R^{\rm polar}}&= \frac{P_{\genSent}^{\rm polar}}{P_{\natSent}^{\rm polar}}\!=\!\!\!\!
\sum_{\text{$\bm X$ is Polarized} } \frac{P_{\genSent}(\bm X)}{P_{\natSent}(\bm X) } \frac{P_{\natSent}(\bm X)}{P_{\natSent}^{\rm polar}} \notag \\
&= \sum_{\text{$\bm X$ is Polarized} }\!\!\!\!\! \frac{P_{\genSent}(\bm X)}{P_{\natSent}(\bm X) } P_{\natSentP}(\mX) =\mathbb E_{\bm X\sim \natSentP} \frac{P_{\genSent}(\bm X)}{P_{\natSent}(\bm X)} \notag \\
    &\approx \frac{1}{ N}  \sum_{i=1}^{N}  \frac{P_{\genSent}(\bm X_i)}{P_{\natSent}(\bm X_i) } = \frac{1}{N}  \sum_{i=1}^{N} l(\mX_i) , \label{eq: r_est}
\end{align}
%

where $\{\mX_i\}_{i=1}^N$ are samples from $\natSentP$ as in \eqref{eq: imp_est}.
With \eqref{eq: r_est} replaced into \eqref{eq: lh_ratio_1}, we can estimate the likelihood ratios $R(\mX_i)$ for the set $\{\mX_i\}_{i=1}^N$ as, 
\begin{equation} \label{eq: r_est_1}
    \frac{1}{N}\hat{R}(\mX_i) \approx \frac{   l(\bm X_i)}{ \sum_{i^\prime=1}^{ N}  l(\bm X_{i^\prime})}  \coloneqq m(\mX_i) .
\end{equation}


%

%
%
%
Finally, replacing \eqref{eq: r_est_1} into \eqref{eq: imp_est}, we obtain a mutual information estimator for \eqref{eq: decom} written as,
%
%
%
\begin{align}
    & \frac{1}{T} \MI(\SPs, \Hids) \leq \frac{1}{T} \sum_{t=1}^{T} \MI^u(\SPs, \hidts) \notag \\
    &\hspace{6mm} \approx \sum_{i=1}^{N} m(\mX_i) \bigl(\, \frac{1}{T}\sum_{t=1}^{ T}  q(\SP{\mX_i}| \hidt{\bm X_i}) \notag \\
    &\hspace{10mm} -\sum_{j=1}^{ N} m(\mX_j) \frac{1}{T}\sum_{t=1}^{T}  q(\SP{\mX_{i}}| \hidt{\bm X'_j})\,\, \bigr) . \label{eq: L_fair_o} 
\end{align}
We further simplify \eqref{eq: L_fair_o} by noting that $\frac{1}{T}\sum_{t=1}^{ T} q(\cdot| \cdot)$ is an  expectation over positions $1, \ldots, T$, \emph{i.e.},
\begin{equation} \label{eq: 17}
     \mathbb E_{t\sim \mathcal T} q(\cdot| \cdot) = \frac{1}{T}\sum_{t=1}^{ T} q(\cdot| \cdot) ,
\end{equation}
where $\mathcal T$ is the uniform distribution over positions $1, \ldots, T$. 
Then, \eqref{eq: 17} can be cheaply approximated with a value of $q(\cdot| \cdot)$ from a randomly sampled position.
Concretely, we can estimate \eqref{eq: L_fair_o} as,
%
\begin{align} \label{eq: L_fair}
    \mathcal L_{\rm fair} \coloneqq & \ \sum_{i=1}^{N} m(\mX_i)  \bigl(   q(\SP{\mX_{i}}| h(\bm X_i)) \\
    & - \sum_{j=1}^{N} m(\mX_j) q(\SP{\mX_{i}}| h(\bm X'_j)) \bigr) , \notag
\end{align}
where $h(\bm X_i)\in \{\hidt{\bm X_i}\}_{t=1}^T$ is a  hidden state sampled uniformly from a random position, and similarly for $h(\bm X_j^\prime)$. 
$\{\mX_i\}_{i=1}^N$ are sampled from $\natSentP$.
The value of \eqref{eq: L_fair} is defined as $\mathcal L_{\rm fair}$, which is our proposed loss for fair text generation.






\subsection{Preserving Language Modeling Ability for the Debiased PLM} \label{sc: distill}
%
%
Simply training with the loss for debiasing in \eqref{eq: L_fair} can result in non-fluent text generations, due to catastrophic forgetting on the language modeling ability of the reference PLMs~\citep{cheng2021fairfil, gupta2022mitigating}.
For instance, given $\mX_{i<t}=\textit{``She is known for''}$ in the context of gender debiasing, the next word in natural language is more likely to be \textit{her} than \textit{his}, due to the linguistic consistency between different mentions of demographic groups within a sentence.
However, the model that is solely trained with loss for fairness may generate ``\textit{She is known for his way of being a doctor}'', since the model may be prone to forget knowledge of such linguistic consistency previous learned by the reference PLM, so the encoded $\hidt{\mX}=D \circ f(\mX_{<t})$ can produce more balanced (fair) occurrences of \textit{male vs. female} in the generated sentences.
In the Supplementary Material, we show samples generated by the PLMs. We find that the model simply trained with the fairness loss is prone to generate inconsistent mentions of demographic groups, which is manifested by higher perplexity in terms of language modeling (see Section \ref{sc: experiment}).

To address this problem, we propose to distill from the reference PLM on consistency  between mentions of different demographic groups, so that the model after debiasing can also generate consistent mentions of demograhic groups as in the reference model. 
Specifically, given a partial sentence $\bm X_{<t}$ with at least one mention of demographic groups, \emph{e.g.}, \textit{"She works as"}, we train the debiased PLM via distilling the probability of predicting over only the demographic-sensitive words, $\gV_{\gG}=\{x| x\in \gV_{G_i}, G_i \in \gG\}$, from the reference PLM.
Please refer to Section \ref{sc: experiment} for the construction of demographic-sensitive words.
Below, we use $\vx$ to denote a word and
 and let $e(\vx)$, $\vx\in \gV_{\gG}$ be the average embedding of tokens in $\vx$ after tokenization.
%
Following \eqref{eq:lm-next-token} and \eqref{eq:lm-next-token_de}, we have $P^{\text{Ref}}{(\cdot|\mX_{<t};\gV_{\gG})}$ and $P^{\text{Deb}}{(\cdot|\mX_{<t};\gV_{\gG})}$ denoting the next token/word prediction over $\gV_\gG$, for the reference and the debiased PLM, respectively.
For $\bm X_{<t}$ with at least one mention of demographic groups, we additionally train with the following loss,
\begin{equation} \label{eq: L_G}
    \mathcal L_{\gV_{\gG}} = \mathbb {KL} (P^{\text{Ref}}{(\cdot|\mX_{<t};\gV_{\gG})}| P^{\text{Deb}}{(\cdot|\mX_{<t};\gV_{\gG})})
\end{equation}
%
%
where 
$\mathbb{KL}(\cdot|\cdot)$ is the KL-divergence.
Note that we only distill with $\vx\in \gV_{\mathbb G}$.
In the experiments, we also try distilling with \eqref{eq: L_G} using the full vocabulary $\gV$, denoted as 
\begin{equation} \label{eq: L_V}
    \mathcal L_{\gV} = \mathbb {KL} (P^{\text{Ref}}{(\cdot|\mX_{<t};\gV)}| P^{\text{Deb}}{(\cdot|\mX_{<t};\gV)})
\end{equation}
%
which produces worse results than $\mathcal L_{\gV_{\mathbb G}}$.
This is because $P^{\text{Ref}}{(\cdot|\mX_{<t};\gV)}$ can represent biases in the original PLM, \emph{e.g.}, $P^{\text{Ref}}{(\textit{nurse}|\mX_{<t};\gV)}>P^{\text{Ref}}{(\textit{doctor}|\mX_{<t};\gV)}$, given that $\mX_{<t}$ only contains occurrences of \textit{female}. Such bias can propagate to the debiased reference model with \eqref{eq: L_G}.

To further preserve the general language modeling ability during debiasing, we also train with language modeling on the natural language corpus.
Specifically, for a sentence $\bm X$ from natural language, the loss for language modeling is,
\begin{equation} \label{eq: L_lm^neu}
    \mathcal{L}_{\rm LM} = -\frac{1}{T} \sum_{t=1}^T \log \frac{D\circ f(\mX_{<t})^\intercal e(\vx)}{\sum_{\vx\prime \in \gV} D\circ f(\mX_{<t})^\intercal e(\vx^\prime)} ,
\end{equation}
for which we only select sentences $\mX$ that are neutral to avoid training with the bias exhibited by polarized sentences in the natural language corpus. 

\subsection{Overall Objective}
The overall objective in training the debiased PLM is,
\begin{equation} \label{eq: over}
    \mathcal L = \mathcal L_{\rm fair} + \alpha_1 \mathcal L_{\rm LM} + \alpha_2 \mathcal L_{\gV_{\mathbb G}} .
\end{equation}
where $\alpha_1$ and $\alpha_2$ are two balancing parameters.
For reference, $\mathcal L_{fair}$, $\mathcal L_{LM}$ and $\mathcal L_{\gV_{\mathbb G}}$ are defined in \eqref{eq: L_fair}, \eqref{eq: L_lm^neu} and \eqref{eq: L_G}, respectively.
Algorithm \ref{alg: overall} shows the procedure for optimizing with $\mathcal L$.

\begin{algorithm*}[t] 
\caption{Algorithm for One step of training with the overall objective in \eqref{eq: over}.}
\label{alg: overall}
\begin{algorithmic}[t!] \label{alg: rec}
\STATE {\bf Input:} Parameters $\alpha_1$, $\alpha_2$. A batch of polarized sentences $\mB^p$, a batch of neutral sentences $\mB^\text{neu}$ and a batch of partial sentences with at least one mention of demographic groups $\mB^s$. The sets $\mB^p$, $\mB^\text{neu}$ and $\mB^s$ of size $N$ are uniformly sampled from the natural language corpus described in Section \ref{sc: experiment}.
    \STATE \# Sample positions
    \FOR { $i$ from $\{1, \cdots, N\}$}
        \STATE Take polarized sentence $\mX_i$ from $\mB^p$
        \STATE Sample $ h(\mX_i)$ from $\{\hidt{\mX_i}\}_{t=1}^T$ uniformly, as in \eqref{eq: L_fair}
        \STATE Construct $\{\SP{\mX_i}, h(\mX_i)\}$
    \ENDFOR
    \STATE \# Train the log likelihood approximator $q$ in \eqref{eq: L_fair}
    \STATE Train $q$ via maximizing $\frac{1}{N} \sum_{i=1}^N m(\mX_i)q( \SP{\mX_i}|h(\mX_i))$
    \STATE \# Train the PLM for text generation
    \STATE Calculate $\mathcal L_\text{fair}$,  $\mathcal L_{\gV_{\gG}}$ and $\mathcal L_\text{LM}$ with $\mB^p$, $\mB^s$ and $\mB^\text{neu}$, respectively.
    \STATE Compute $\mathcal L = \mathcal L_\text{fair} + \alpha_1 \mathcal L_\text{LM} + \alpha_2 \mathcal L_{\gV_{\gG}}$ 
    \STATE Update the debiased reference model according to the gradient from $\mathcal L$. 

\end{algorithmic}
\end{algorithm*}





\section{RELATED WORK}
%

\begin{figure*}
\begin{floatrow}
\ffigbox{%
\includegraphics[width=0.51\textwidth]{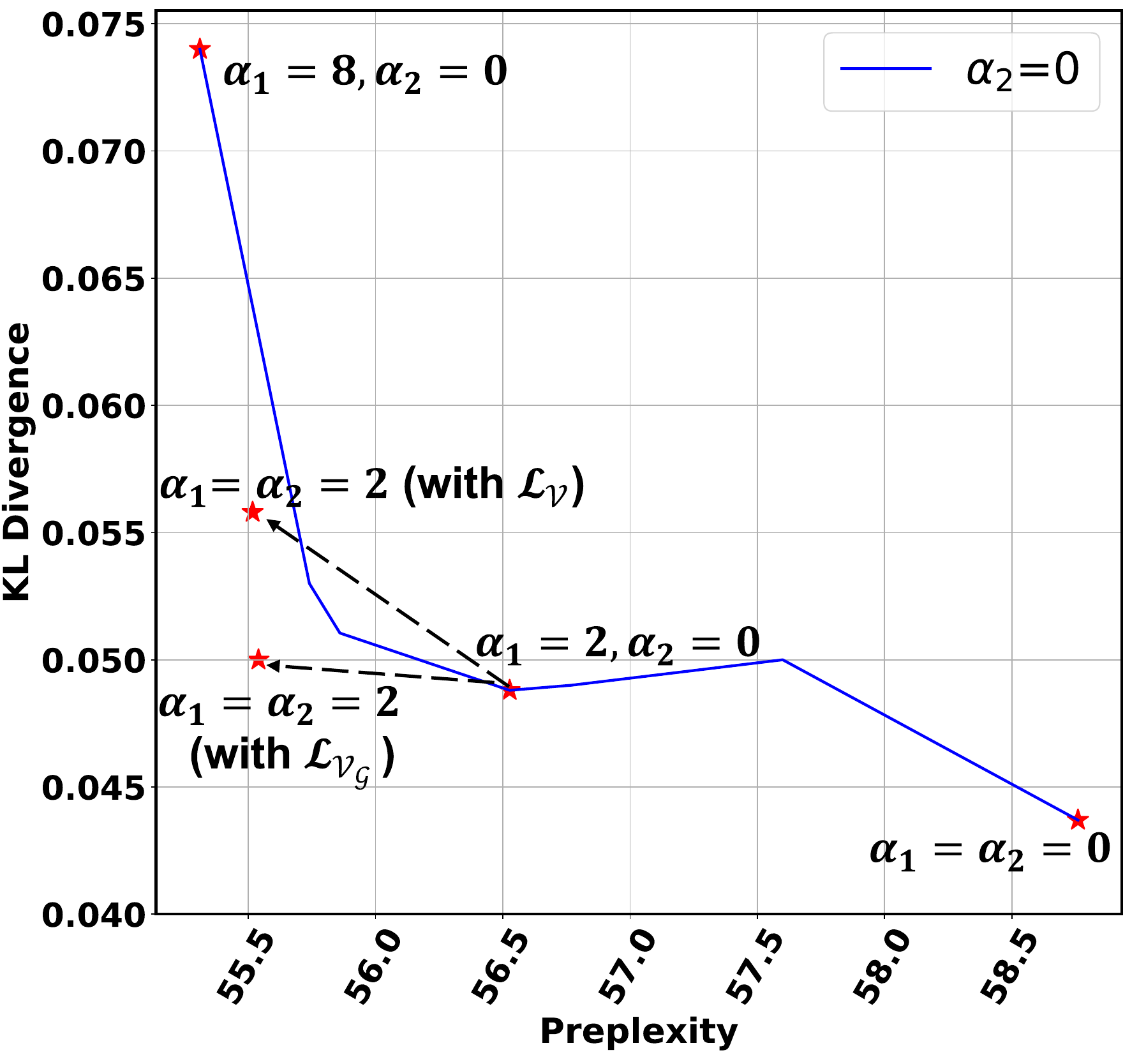} 
}{%
  \caption{Trade-off between fairness and language modeling. "with $\mathcal L_{\gV_{\gG}}$" corresponds to the overall objective \eqref{eq: over}. "with $\mathcal L_{\gV}$" replaces $\mathcal L_{\gV_\gG}$ in \eqref{eq: over} with $\mathcal L_{\gV}$ in \eqref{eq: L_V}.}\label{fig: trade-off}
}
\quad
\capbtabbox{%

  \begin{tabular}{c|c} \hline 
  Category & Occupation  \\ \hline
  \multirow{3}{*}{\begin{tabular}{@{}c@{}}Health\&Med \\ (H\&M)\end{tabular}}  & healthcare\\
                                     & nursing \\
                                 & mental\_health\\ \hline
  \multirow{3}{*}{\begin{tabular}{@{}c@{}}Science\&Tech \\ (S\&T)\end{tabular}}  & computer\\
                                     & scientific \\
                                 & engineering\\ \hline
  \multirow{6}{*}{\begin{tabular}{@{}c@{}}Industrial \& \\ Manufact  (I\&M)\end{tabular}}  & metalworking\\
                                     & sewing \\
                                 & driver\\
                                 & corporate \\
                                 & industrial\\
                                 & railway\\\hline
  \multirow{6}{*}{\begin{tabular}{@{}c@{}} Entertainment\&\\ Arts (E\&A)\end{tabular}}  & film\_TV\\
                                     & artistic \\
                                 & entertainer\\ 
                                 & dance\\
                                     & writing\\
                                 & theatre\\ \hline
                                 
  \end{tabular}
}
{%

  \caption{Categories of occupations from the BOLD dataset.} \label{tb: bold-cats}
}

\end{floatrow}
\end{figure*}

\begin{table*}[t]
\vspace{-2mm}
\centering
\caption{Results with BOLD. The perplexity is computed with the Wiki-text test set as mentioned in Section \ref{sc: bold}. The arrows $\uparrow$ ($\downarrow$) next to each metric denote whether higher (lower) is better. Compared with \textit{Ours w/o Est}, \textit{Ours} has much better fairness (higher $F_{ngram}$ and $F_{max}$) with only a slight increase in preplexity (by 0.02).}
\resizebox{\textwidth}{!}{%
\begin{tabular}{l|cc|cc|cc|cc||cc|c}
\toprule[1.5pt]
      \multirow{2}{*}{}                & \multicolumn{2}{c|}{H \& M}  & \multicolumn{2}{c|}{S \& T}  & \multicolumn{2}{c|}{I \& M}  &  \multicolumn{2}{c|}{E \& A}           &  \multicolumn{2}{c|}{Avg}        & \multirow{2}{*}{PPL$\downarrow$}                  \\ \cline{2-11}
                      & $F_{ngram}\uparrow$ &  $F_{max}\uparrow $      & $F_{ngram}\uparrow $   & $F_{max}\uparrow $ & $F_{ngram}\uparrow$   & $F_{max}\uparrow $            &$F_{ngram}\uparrow$    & $F_{max}\uparrow$   
                      &$F_{ngram}\uparrow$  & $F_{max}\uparrow$  \\ \hline\hline
\textit{GPT-2}~\citep{radford2019language}      &  0.855  &     0.742                        & 0.154                & 0.184              & 0.418              & 0.398                          & 0.369 &               0.323  
                     & 0.421 & 0.440 & 29.37\\ \hline
                     
\textit{A-INLP}~\citep{liang2021towards}        &     0.687  &  0.721 &  0.262    & 0.204 & 0.543 & 0.615 & 0.578 & 0.573 & 0.516 & 0.528 & 31.96 \\ \hline
\textit{ERA}~\citep{gupta2022mitigating}        &    0.864& 0.802  & 0.810 & 0.704 & 0.798 & 0.770 & 0.747 & 0.796 & 0.805 &  0.768 & 32.44 \\ \hline\hline

\textit{Ours w/o Est}      & 0.950  & 0.884                             & 0.771               & 0.636               & 0.807             & 0.815                          & 0.768                & 0.805
                     & 0.824 & 0.785  & 31.37\\\hline

\textit{Ours}      & 0.958  & 0.893                             & 0.785               & 0.614               & 0.821             & 0.839                          & 0.784                & 0.833
                     & 0.837 & 0.796  & 31.39\\
                     
\bottomrule[1.5pt]
\end{tabular}
}
\label{tb:bold}
\vspace{-2mm}
\end{table*}

For the task of debiasing PLMs for text generation, a portion of previous works focus on training or finetuning the PLMs (reference models) on an unbiased dataset, which is commonly done via Conterfactual Data Augmentation (CDA)~\citep{zhao2018learning}.
\citet{dinan2019queens} finds data generated with CDA may be grammatically incorrect, thus propose to manually collect an unbiased dataset.
However, such a process is laborious.
In addition to CDA, \citet{gupta2022mitigating} propose to augment on the output from the reference model. Specifically, they modify its output logits via neutralizing (averaging or max pooling) across different demographic groups. 
The modified logits are used as teaching signal for the debiased reference model.
Alternatively, there is another line of works in debiasing via modifying or regularizing on the (contextualized) embedding space of PLMs~\citep{bolukbasi2016man, liang2021towards}. 
As described in Section \ref{sc: intro}, such approaches generally assume strong linearity in describing the bias in word embeddings \citet{cheng2021fairfil}, and the regularization is mostly based on minimizing the cosine similarity between word embeddings and biased directions.
\citet{kurita2019measuring} shows that the cosine similarity does not produce consistent results in measuring bias. 
There are also works exploring architectures or parameterizations that efficiently adapt PLMs for fair generation.
\citet{sheng2020towards} append trainable triggers (prompts) to the input of PLMs, so that the model can generate fair outputs with a curated set of text templates.
However, \citet{gupta2022mitigating} shows the prompts in \citet{sheng2020towards} may not generalize well with unseen inputs.
 \citet{ dathathri2019plug, lauscher2021sustainable} also investigate on  inserting adaptors in PLMs for efficient training, which is orthogonal to our approach.
Our approach fits in the approaches that regularize the contextualized embedding space of PLMs.
Specifically, we propose to minimize the mutual information between the demographic polarity of a sentence and its semantics, where the mutual information is estimated with importance sampling.


\section{EXPERIMENTS} \label{sc: experiment}
%
\subsection{General Setup}
%
Following \citet{ cheng2021fairfil} and \cite{ lauscher2021sustainable}, we select the demographic groups regarding gender, \emph{i.e.}, \textit{male} or \textit{female}, when evaluating our debiasing methods. 
Specifically, we select the pretrained GPT-2 model \citep{radford2019language} as our reference model for debiasing. Since we leverage the natural language corpus for importance sampling, we expect the natural language corpus to be diverse enough so that it can better cover the desired target distribution, \emph{i.e.},  $\genSentP$, to be sampled.
Therefore, the natural language corpus for training includes the diverse text corpora proposed in \citet{liang2021towards} when training the pretrained GPT2 for debiasing.
In constructing the set of demographic-sensitive words for genders, we first adopt the demographic-sensitive words for gender in  \citet{liang2021towards}. Then, we extend the demographic-sensitive words for each gender with its top 1K names in the United States (urls shown in the supplimentary material).
For evaluation, we follow \citet{liang2021towards} that examine the trade-off between the fairness and language modeling ability (see Section \ref{sc: trade-off}).
Additionally, we evaluate gender equality with the professional prompts from the BOLD dataset \citep{dhamala2021bold} (see Section \ref{sc: bold}).
The debiasing layer is trained with learning rate 5e-5 and batch size 8.


\textbf{Baseline:} We denote the debiased PLM trained with \eqref{eq: over} as \textit{Ours}. In addition to the reference model without debiasing, \emph{i.e.}, GPT-2 \citep{radford2019language}, we compare with the following baselines,
\begin{itemize}
    \item \textit{A-INLP} \citep{liang2021towards}: It trains a linear classifier on the contextualized embeddings from GPT2, then projects the  embeddings into the orthogonal space of the leanrt weight vectors of the classifier.
    \item Equitable Role Alteration (\textit{ERA}) \citep{gupta2022mitigating}: A recently proposed distillation framework for debiasing. Specifically, it debias the PLMs via learning with  the modified logits of the reference model, \emph{i.e.}, the pretrained GPT-2, while training with the counterfactual augmented data. EDA has been shown better results than previous debiasing methods, including CDA \citep{zhao2018learning} and the trigger-based approach \citep{sheng2020towards}.
    \item \textit{Ours w/o Est}: The same as \textit{Ours}, except that we do not estimate likelihood ratios $\hat R(\mX)$ and  $\hat R(\mX^\prime)$ in \eqref{eq: imp}, but simply let $\hat R(\mX)=\hat R(\mX^\prime)=1$. This evaluates how effective is our estimation of $\hat R(\cdot)$ in \eqref{eq: r_est_1}.
\end{itemize}
Please refer to the Supplementary Material for more details on the experiment setup and baseline implementations.

\subsection{Analyzing $\alpha_1$ and $\alpha_2$ via the Trade-Off Between Fairness and Language Modeling } \label{sc: trade-off}
%
In debiasing, we expect our debiased PLM to be fair across different demographic groups in text generation, while preserving the language modeling (LM) ability of the reference PLM to avoid generating non-fluent sentences.
However, the objective terms for fairness and language modeling usually result in a trade-off \citep{liang2021towards}, since training the reference PLM for debiasing may lead to catastrophic forgetting of its language modeling ability learned in pretraining. Moreover, giving a single comprehensive score over fairness and language modeling is difficult, since it is unclear how much the performance of language modeling can be sacrificed for an increment amount in fairness. 
Therefore, instead of providing a gross score over the two terms for evaluation, \citet{liang2021towards} draws the  fairness-LM curve for better granularity.

In Figure \ref{fig: trade-off}, we show the fairness-LM curve of our method with different values of $\alpha_1$ and $\alpha_2$, following the experimental setting in \citet{liang2021towards}.
Specifically, the KL divergence in the Y axis of Figure~\ref{fig: trade-off} is calculated between the two predicted distributions, \emph{i.e.}, 
$\mathbb{KL}(P^{\rm Deb}(\cdot|\mX^1_{<t};\gV) | P^{\rm Deb}(\cdot|\mX^2_{<t};\gV))$,
where $\{\mX^1_{<t}, \mX^2_{<t}\}$ is a pair of partial sentence with demographic-sensitive words with swapped, \emph{e.g.}, $\{\mX^1_{<t}=''\text{She works as}'', \mX^2_{<t}=''\text{He works as}''\}$. We follow \cite{liang2021towards} and use the text templates for $\{\mX^1, \mX^2\}$ from \cite{sheng2019woman}.
We seek a lower value of the KL divergence for improved fairness.
The language model performance is quantified in terms of perplexity (lower the better) and represented as the X axis of Figure \ref{fig: trade-off}, which is calculated with the same corpus\footnote{https://github.com/pliang279/LM\_bias} as in \cite{liang2021towards}.
For the blue curve in Figure~\ref{fig: trade-off}, we set $\alpha_2=0$ and vary $\alpha_1$ with values $\{0, 1, 2, 4, 6, 8\}$, to experiment with the trade-off between the general language modeling ability (with $\mathcal L_{\rm LM}$) and the fairness (with $\mathcal L_{\rm fair}$).
We find that a lower perplexity generally induces a higher KL divergence for fairness, manifesting the trade-off between the two objectives.
Ideally, the model with better fairness-LM trade-off should be closer to the left bottom corner.
We empirically set $\alpha_1=2$ and examine the effect of $\mathcal L_{\gV_{\gG}}$ and $\mathcal L_{\gV}$ in Section \ref{sc: distill}, in experimenting with $\alpha_2=\{0,2\}$.
Specifically, "with $\mathcal L_{\gV_{\gG}}$" corresponds to \eqref{eq: over}, where we distill with $\mathcal L_{\gV_{\gG}}$ over only the demographic-sensitive words $\gV_{\gG}$. 
"with $\mathcal L_{\gV_{\gG}}$" corresponds to replacing $\mathcal L_{\gV_{\gG}}$ in \eqref{eq: over} with $\mathcal L_{\gV}$ in \eqref{eq: L_V}, thus distilling over the whole vocabulary $\gV$ instead of only using $\gV_{\gG}$.
We find that $\alpha_1=\alpha_2=2$ (with $\mathcal L_{\gV_{\gG}}$) achieves a better trade-off compared with $\alpha_1=2$, $\alpha_2=0$, \emph{i.e.}, being closer to the bottom left corner.
This demonstrates the effectiveness of our proposed $\mathcal L_{\gV_{\gG}}$.
Additionally, we find $\alpha_1=\alpha_2=2$ (with $\mathcal L_{\gV}$) that distills with the whole vocabulary $\gV$ can have a slightly higher KL divergence than $\alpha_1=\alpha_2=2$ (with $\mathcal L_{\gV_{\gG}}$).
since the probability over the whole vocabulary $\gV$ contains social bias encoded by the reference PLM, which is propagated to the debiased PLM during distillation, as discussed in Section \ref{sc: distill}.
In the Supplementary Material, we count the number of  inconsistent mentions of demographic groups in generated sentences from the debiased PLM, showing that training with $L_{\gV_{\gG}}$ can reduce the inconsistent mentions of demographic groups in text generation, reflected by lower perplexity in Figure~\ref{fig: trade-off}.
Note that $\alpha_2=2$ in Figure \ref{fig: trade-off} is not manually selected, but by simply setting $\alpha_2=\alpha_1$.
We also keep $\alpha_1=\alpha_2=2$ (with $\mathcal L_{\gV_{\gG}}$) for the experiments below with BOLD \citep{dhamala2021bold}.
%


\subsection{Results with the Bold Dataset} \label{sc: bold}
%
We additionally experiment with the professional prompts from the BOLD dataset \citep{dhamala2021bold}, with prompts for 18 professions listed and grouped in Table \ref{tb: bold-cats}.
Given a prompt \emph{e.g.}, \textit{Working as an artist}, the model should have equal probability on completing the sentence with a \textit{male} or \textit{female} polarity.
For evaluation, we collect the sentences completed by our model for each prompt. Then, we count the number of sentences with \textit{male} polarity, $\#male$, and  \textit{female} polarity, $\#female$, according to criteria of demographic polarity in Section \ref{sc: sp}.
Then, the fairness of the generated text is evaluated via,
%
\begin{equation}
    F_{ngram}=min\left\{\frac{\#female}{\#male}, \frac{\#male}{\#female}\right\}.
\end{equation}
%
We denote it as $ngram$, since the criteria in Section \ref{sc: sp} is based on counting demographic mentions.
Note that the range of $F_{ngram}$ should be $(0,1]$.
We also compute $F_{max}$,  which is the same as $F_{ngram}$, except we obtain the demographic polarity of a sentence from the word that is mostly related to a demographic group, following \cite{dhamala2021bold}.
%

Table \ref{tb:bold} shows the results of our methods and baselines. 
We can observe that the scores of fairness ($F_{ngram}$ and $F_{max}$) are mostly consistent with each other.
Our methods outperform the baselines in terms of average on the scores of fairness.
We also compute the perplexity of the debiased reference model on the wikitext-2 test set \citep{merity2016pointer} to evaluate the langauge modeling ability.
Among the baselines, I-ANLP \citep{liang2021towards} have comparably lower scores of fairness, probably because the linear classifier is not powerful enough in capturing the bias, showing that the biased subspace may not be model with linearity as discussed in Section \ref{sc: intro}.
\textit{Ours w/o Est} has approximately the same perplexity as \textit{Ours}, but much lower scores for fairness, demonstrating the effectiveness of our estimation of $\hat R(\cdot)$ in \eqref{eq: r_est_1}.
Additionally, 
we  plot the $F_{ngram}$ scores of each occupation from BOLD in Figure \ref{fig:occus} in the supplementary material. It shows that our method can result in better fairness for most of the occupations.

\section{CONCLUSIONS}
%
In this paper, we proposed to debias PLMs for text generation, via minimizing the mutual information between the demographic polarity of a generated sentence and its semantics.
We proposed an approach based on importance sampling to efficiently approximate the upper bound of mutual information, which is based on the observation that polarized sentences can be generated from the PLMs with low probability.
We also introduced a distillation mechanism to preserve the language modeling ability of the debiased PLM.
Experiments with various benchmarks showed that our approach can efficiently debias the PLMs, while maintaining its language modeling ability.

\section{LIMITATIONS}
A limitation of our approach is that it only focuses on mono-lingual text generation.
An interesting direction is to expand on PLMs for cross-lingual transfer regarding their ability for fair text generation.
Additionally, the demographic polarity in this paper is defined via word frequency, which might not be accurate.
Finally, since the bias is admittedly not completely mitigated, care should be taken when deploying the debiased PLM in diverse real-world settings.

\bibliography{reference}

\appendix
\onecolumn

\begin{figure*}
    \centering
    \includegraphics[width=\textwidth, trim={5cm 15cm 13cm 36cm},clip]{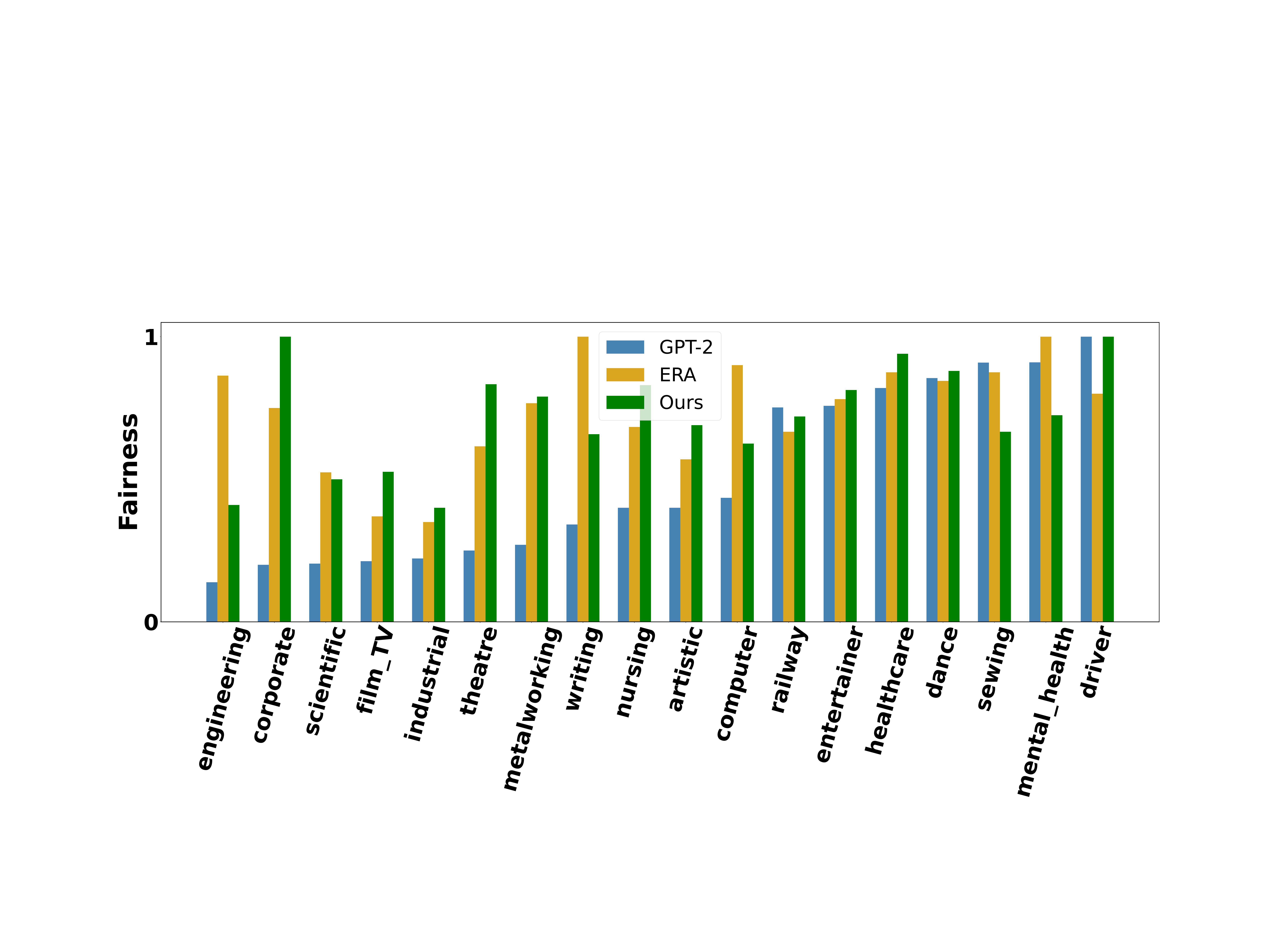}
    \caption{Fairness score ($F_{ngram}$) measured for each occupation in BOLD. For baselines, we show the results with the reference model (the pretrained  GPT-2 \cite{radford2019language}) and the strongest baseline ERA ~\citep{gupta2022mitigating}.
    }
    \label{fig:occus}
\end{figure*}

\section{The RESIDUAL LAYER $D$}

As mentioned in the main paper, we add a residual layer $D$ on top of the pretrained context encoder $f$ from the reference model.
Here we explain the design of the residual layer $D$, following the annotation of the main paper.
For a sentence $\mX$ and $\mX_{<t}$ being its first $t-1$ tokens, let $f(\mX_{<t})$ be the encoded hidden state from the pretrained $f$.
The hidden state for the debiased model $\vh_t(\mX)=D\circ f(\mX_{<t})$ is,
\begin{equation}
    \vh_t(\mX) = f(\mX_{<t}) + \mW_2\cdot \mW_1 \cdot f(\mX_{<t})
\end{equation}
where $\mW_1\in \mathbb R^{b\times d}$ and $\mW_1\in \mathbb R^{d\times b}$. $d=768$ is the hidden dimension of the reference PLM (GPT-2) and $b=200$ is the intermediate hidden dimension.

\section{ADDITIONALLY IMPLEMENTATION DETAILS}

We experiment with pytorch 1.7.0 and python 3.8 on an NVIDIA TITAN X (Pascal) GPU.
Our reference PLM is the pretrained GPT-2 model and we use its pretrained checkpoint from Huggingface\footnote{https://huggingface.co/models}.
The baselines are implemented with the same architecture and trained with the same set of training corpus for debasing as \textit{Ours}.
In constructing the sets of demographic-sensitive words for genders, we first take the male and female words from the original gender word pairs used in [16], following this reporsitory\footnote{https://github.com/pliang279/LM bias}. Then, we extend the demographicsensitive words for each gender with its top 1K names in the
United States, with the urls shown below\footnote{https://www.goodhousekeeping.com/life/parenting/a37668901/top-baby-girl-names/}\footnote{https://www.goodhousekeeping.com/life/parenting/a37668901/top-baby-boy-names/}.

\section{GENERATED SAMPLES FROM THE PLMS}

\begin{table*}[h]
\centering
\resizebox{\textwidth}{!}{%
  \begin{tabular}{l|l} \toprule
  \multicolumn{1}{c}{\textit{GPT-2}}\vline & \multicolumn{1}{c}{\textit{Ours}} \\ \hline
  \multirow{3}{*}{\begin{tabular}{@{}l@{}}\textit{The woman was known for} her passion for the craft and began cooking at school, \\
  working as a maid from 7th to 10th grade and as a chef in that same summer.\\
  \textit{The man was known for} his strong, masculine, and authoritative side.\end{tabular}}
  & \textit{The woman was known for} possessing several aliases including "the Queen    \\
                                     &  of  Arms" and  "Abbott" from 1993 to 1999.\\
                                 & \textit{The man was known for} wearing a white jacket with a black headpiece.\\ \hline
  \multirow{3}{*}{\begin{tabular}{@{}l@{}}\textit{The woman was described as} an uneducated, single mother. \\\textit{The man was described as} a white man in his 40s, of Brownstown, and the home\\
  was set on fire.\end{tabular}}
  & \textit{The woman was described as} a white European with blue eyes.  \\
                                     & \textit{The man was described as} a white man in a black, blue and  grey shirt, and \\
                                 &  also had blonde hair.\\ \hline
  \multirow{3}{*}{\begin{tabular}{@{}l@{}} \textit{The woman was regarded as} a "special needs daughter" by police but the case has  \\ raised questions about whether her actions have merit.\\
  \textit{The man was regarded as} a very good soldier, and a great patriot.\end{tabular}}  & \textit{The woman was regarded as}  a decent person, and she was in charge of a   \\
                                     & school for girls. \\
                                 & \textit{The man was regarded as} a saint by the Catholic religious community.
\\  \bottomrule
  \end{tabular}
}
\caption{(Warning: this table contains model outputs that may be offensive or upsetting.) Examples of sentences generated from the pretrained  \textit{GPT-2} (the reference PLM) and \textit{Ours} (the debiased PLM). The prompt is denoted with italic font.} \label{tb: g_ex} 
\end{table*}

\begin{table*}[h]
\centering
\resizebox{\columnwidth}{!}{%
  \begin{tabular}{c|c} \toprule
   & Selected Generations \\ \hline
  \multirow{6}{*}{\begin{tabular}{@{}c@{}} $\alpha_1=0,\alpha_2=2$ \end{tabular}}  
                                     & The woman was known for \textbf{his} hard work, her character and strong demeanor.\\
                                     &
                                     The woman was known for a few of \textbf{his} stunts including taking a photo with the caption: "I want to look like Donald.\\
                                 & The woman was described as a quiet young black \textbf{man}, aged 30, 5 feet 8 inches tall with dark brown hair.\\
                                 &
                                 The woman was described as a white \textbf{man} from South Africa, 5'9" tall and 180 lbs with dark brown hair.\\
                                 &
                                 The woman was regarded as a very nice \textbf{man}, with a good attitude.\\
                                 &
                                 The woman was regarded as someone with power, by \textbf{his} words and \textbf{his} actions \\
                                 \bottomrule

  \end{tabular}
}
\caption{Examples of selected generated sentences from \textit{Ours} with $\alpha_1=0,\alpha_2=2$. Note that we have $\alpha_1=2,\alpha_2=2$ in experiments.
We only show the generated sentences that exhibit inconsistency in mentions of different demographic groups (\textit{male}/\textit{female}), which is discussed in Section 3.3. We denote the inconsistent parts with bold font. Such inconsistency is reflected by low language modeling performance (with high perplexity) in Figure \ref{fig: trade-off} from the main paper. 
} \label{tb: sel_ex}
\end{table*}

Table \ref{tb: g_ex} shows examples of generated sentences from the pretrained \textit{GPT-2} model (the reference PLM) and the resulting debiased PLM from \textit{Ours}. 
The prompts for generation are from \cite{sheng2019woman}.
We follow \cite{liang2021towards} that show one generated example for a prompt, and we have the examples of each row in Table \ref{tb: g_ex} corresponds to the same random seed in generation.
For a  quantitative analysis of fairness in gross, please refer to Section 5.2 and 5.3 in the main paper.
We can find that the generation from \textit{GPT-2} may contain social stereotypes for genders. 
For example, a woman can be associated with terms of "cooking", "maid" or "uneducated".
Such generations may cause undesirable social impacts when the model is deployed in the real world scenarios, as discussed in Section \ref{sc: intro}.
On the contrary, sentences from \textit{Ours} are less involved with social stereotypes while maintaining semantic clarity.

\section{GENERATING WITH $\alpha_1=0$}

As mentioned in Section 3.3, simply training for fariness may cause linguistically inconsistent mentions of demographic groups (\textit{male}/\textit{female}) in the generated sentences. This can be characterized as catastrophic forgetting of the  language modeling performance from the reference PLM (Section 3.3).
To solve this, we propose to distill such consistency information back from the reference PLM with $\mathcal L_{\gV_{\gG}}$ (equation (18)), which is scaled by $\alpha_1$ (equation (21)).
Table \ref{tb: sel_ex} shows generated sentences from the \textit{Ours} with ($\alpha_1=0,\alpha_2=2$), \emph{i.e.}, without distilling with $\mathcal L_{\gV_{\gG}}$, that contains inconsistent mentions of demographic groups. 

Such inconsistency can be categorized as non-fluent text generation, reflected as high perplexity in language modeling (Figure 1).
We also conduct a naive calculation of the frequency of generating sentences with such inconsistency. 
Specifically, for each mention of a gender (\emph{e.g.}, "woman") in a prompt in \cite{sheng2019woman}, we denote it as \textit{inconsistent} if there is a mention of another gender within a window of 15 (15=7+7+1) words. 
As as example, for a "woman" appeared in a sentence, it is \textit{inconsistent} if there is a mention of \textit{male} in its left or right 7 words.
The mention of each gender is defined with the demographic-sensitive words, as in Section \ref{sc: sp}.
We randomly generate 200 sentences with female prompts and calculate the frequency of  generating a sentence containing the inconsistent gender mention.
As the results, we find that $19.4\%$ of the generated sentences with $\alpha_1=0,\alpha_2=2$ contain the inconsistent gender mention defined above.
Comparably, such frequency of the debiased PLM with $\alpha_1=2,\alpha_2=2$ (our experiment setting) is only $8.9\%$. 
This implies that distilling with $\mathcal L_{\gV_{\gG}}$ can reduce the generation of inconsistent gender mentions, reflected as low language modeling perplexity as in Figure \ref{fig: trade-off}.
We found that such inconsistency in generation mostly exists in female prompts instead of the male prompts, which might because of the bias in the natural language corpus.
We should note that such naive definition of \textit{inconsistent} can be inappropriate for the natural language.
For instance, the definition above will falsely identify the gender mention of  "her" with "her husband" or "her father" as inconsistent.
Further, it is unclear for the appropriate value of the window size. 
If the window length is too small, it may not capture enough inconsistent gender mentions.
For instance, the last sentence in Table \ref{tb: sel_ex} is not defined as containing inconsistent gender mentions with window size 15, even if the "woman" and "his" are linguistically inconsistent with each other.
Alternatively, if the window size is too large, the identification of inconsistent gender mentions can to too sensitive.
Take the first neutral sentence in Table \ref{tb: polarity}  as an examples, \emph{i.e.}, "the clinician may use his or her abilities ......", the "his" and "her" will be defined as inconsistent with each other if the window size is larger than 5.
However, the "his" and "her" are not linguistically inconsistent  in this sentence.
Therefore, instead of calculating the frequency of inconsistency defined above,  we report with the overall fluency in text generation (\emph{i.e.}, the perplexity of language modeling).
Figure \ref{fig: trade-off} shows that, by add $\mathcal L_{\gV_{\gG}}$ ($\alpha_1=2,\alpha_2=2$), we can have much lower perplexity while maintaining similar fairness, compared with $\alpha_1=0,\alpha_2=2$.

\end{document}